# A Novel Compression Framework for YOLOv8: Achieving Real-Time Aerial Object Detection on Edge Devices via Structured Pruning and Channel-Wise Distillation

Melika Sabaghian[1[0009-0009-0267-5195]], Mohammad Ali Keyvanrad[1[0000-0002-7654-1001]], Seyyedeh Mahila Moghadami[1]

[1] Faculty of Electrical & Computer Engineering, Malek Ashtar University of Technology, Iran
keyvanrad@mut.ac.ir

**Abstract.** Efficient deployment of deep learning models for aerial object detection on resource-constrained devices requires significant compression without compromising performance. In this study, we propose a novel three-stage compression pipeline for the YOLOv8 object detection model, integrating sparsity-aware training, structured channel pruning, and Channel-Wise Knowledge Distillation (CWD). First, sparsity-aware training introduces dynamic sparsity during model optimization, effectively balancing parameter reduction and detection accuracy. Second, we apply structured channel pruning by leveraging batch normalization scaling factors to eliminate redundant channels, significantly reducing model size and computational complexity. Finally, to mitigate the accuracy drop caused by pruning, we employ CWD to transfer knowledge from the original model, using an adjustable temperature and loss weighting scheme tailored for small and medium object detection. Extensive experiments on the VisDrone dataset demonstrate the effectiveness of our approach across multiple YOLOv8 variants. For YOLOv8m, our method reduces model parameters from 25.85M to 6.85M (a 73.51% reduction), FLOPs from 49.6G to 13.3G, and MACs from 101G to 34.5G, while reducing AP50 by only 2.7%. The resulting compressed model achieves 47.9 AP50 and boosts inference speed from 26 FPS (YOLOv8m baseline) to 45 FPS, enabling real-time deployment on edge devices. We further apply TensorRT as a lightweight optimization step. While this introduces a minor drop in AP50 (from 47.9 to 47.6), it significantly improves inference speed from 45 to 68 FPS, demonstrating the practicality of our approach for high-throughput, resource-constrained scenarios.

To our knowledge, this is the first study to integrate CWD-based knowledge distillation with channel pruning on YOLOv8 for aerial object detection, effectively bridging the gap between state-of-the-art detection performance and real-world deployment needs.

## 1 Introduction

The advancements in deep learning have revolutionized artificial intelligence, enabling high accuracy in fields ranging from industry to medicine. Object detection, a crucial application in computer vision, has seen extensive adoption due to these de-velopments. In aerial imaging, object detection is essential for applications such as disaster management, rescue operations, traffic monitoring, and agricultural assessment. However, aerial object detection faces unique challenges that distinguish it from conventional object detection tasks. These challenges include the detection of small, often distant objects, the management of densely distributed targets, and the difficulty of distinguishing objects from intricate, cluttered backgrounds. Despite the significant progress made in this field, current state-of-the-art models for aerial object detection possess complex architectures with numerous parameters, leading to large model sizes, high computational demands, and significant energy and carbon consumption. This renders deployment on resource-limited hardware—such as edge devices, embedded systems, and mobile platforms—impractical.

Among object detection algorithms, the YOLO (You Only Look Once) series has established itself as a leader in real-time detection, renowned for its speed and accuracy. Recent advancements have culminated in the release of YOLOv6 [1], YOLOv7 [2], and YOLOv8 [3] within a single year, with YOLOv8 emerging as the most advanced iteration. YOLOv8 offers notable improvements in performance, adaptability, and accuracy, making it a strong contender for tackling complex detection tasks, particularly in aerial imaging. However, these benefits come at the cost of increased computational complexity. For example, YOLOv8n, a lightweight variant of YOLOv8, requires approximately 3 million parameters and 8.1 GFLOPs—significantly more than YOLOv5n, which uses 1.7 million parameters and 4.2 GFLOPs. Such increases in resource requirements pose challenges for deployment in constrained environments, particularly where computational, memory, and energy resources are scarce.

Recent efforts in model compression for deep learning have focused on methods such as pruning [4], [5], quantization [6], [7], and, increasingly, knowledge distillation [8], [9], [10] to reduce computational demands while maintaining accuracy. However, most studies have concentrated on classification tasks [11], [12], [13] or two-stage detectors [14], [15], with fewer attempts made to compress advanced one-stage models like YOLOv8 for efficient object detection, another significant drawback is that many models compressed using pruning and quantization techniques require specialized hardware or software for deployment, which can raise overall costs.

This gap highlights the need for adaptable and efficient compression methods for single-stage object detectors like YOLOv8, especially in aerial imaging where resource limitations are a significant barrier.

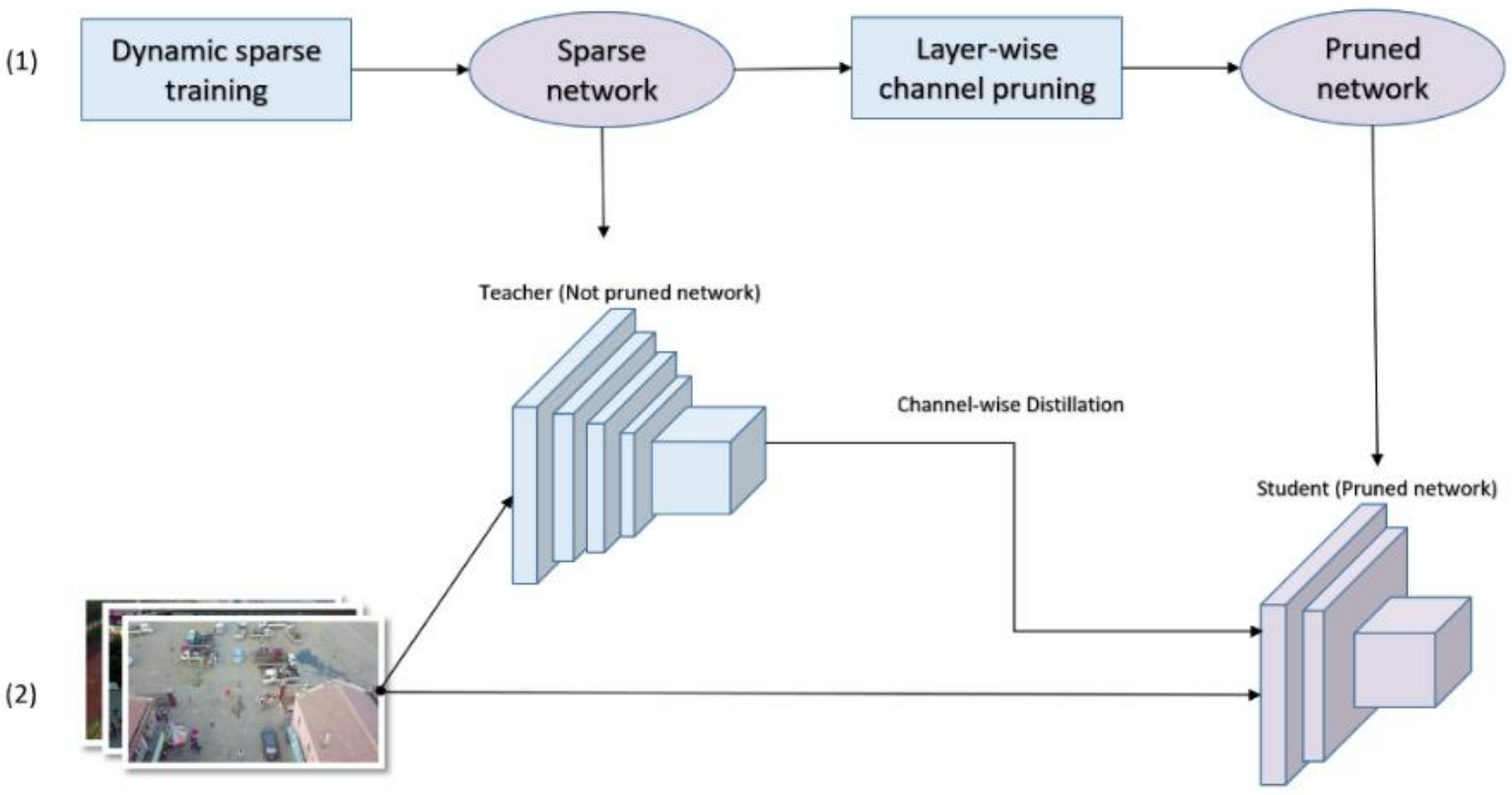

**Fig. 1.** Overview of proposed methods to compress YOLOv8 object detection model to reduce model size, FLOPs and MACs while maintaining the accuracy

To address these challenges, we propose a novel, three-stage compression framework specifically designed to create an efficient and accurate YOLOv8 model for aerial object detection. Our framework as shown in **Fig. 1** operates as follows:

First, we employ Dynamic Sparse Training using L1-norm regularization. This method dynamically adjusts the sparsity rate during training, enabling a gradual and effective induction of sparsity that prepares the model for effective pruning.

Second, we apply a Layer-Wise Structured Pruning algorithm. Unlike traditional methods that use a single global threshold, our approach prunes each layer independently. This fine-grained strategy maximizes the compression potential of robust layers while preserving the integrity of more sensitive ones.

Finally, to recover the performance of the pruned model, we utilize Channel-Wise Feature Distillation (CWD). This distillation method, focused on the critical C2f layers of YOLOv8, transfers essential feature information from the original model, effectively mitigating the accuracy loss caused by pruning.

In summary, the main contributions of this paper are as follows:

- A Novel Compression Framework: We propose an end-to-end framework that synergistically combines dynamic sparse training, layer-wise pruning, and targeted knowledge distillation to significantly compress YOLOv8.
- A Fine-Grained Pruning Strategy: We introduce a layer-wise pruning method that adapts to the specific sparsity distribution of each layer, leading to higher compression rates without the risk of catastrophic layer removal common in global pruning approaches.
- Effective Performance Recovery: We demonstrate the targeted application of Channel-Wise Distillation on the C2f layers of YOLOv8 as an effective technique to restore detection accuracy in a heavily pruned model.
- Extensive Empirical Validation: We provide a thorough evaluation of our framework on the challenging VisDrone aerial imagery dataset, demonstrating signifi-cant reductions in model size, parameters and computational cost (FLOPs) while maintaining competitive accuracy and achieving practical speed-ups on resource constraint hardware.

## 2 Background and motivation

Object detection is a fundamental and widely applied discipline within computer vision and artificial intelligence that has seen a surge in prominence in recent years. Its extensive range of applications has established it as a cornerstone technology in numerous fields. Over the past decade, the advent of convolutional neural networks (CNNs) has driven remarkable advancements in this domain, leading to models with state-of-the-art accuracy. However, the high computational and memory requirements of these models present a significant challenge for their deployment on resource-constrained devices. To address this limitation, researchers have developed various model compression and optimization techniques, with network pruning and knowledge distillation being among the most prominent. **Table 1** provides a comprehensive summary of recent works in this area, outlining their core strengths and limitations.

**Table 1.** Comprehensive summary backgrounds and motivations

| Category | Sub-type / Technique | Representative Examples | Representative Works | Strengths | Limitations |
|---|---|---|---|---|---|
| Object Detection | Two-stage Detectors | R-CNN, Faster R-CNN, Mask R-CNN | [16], [17], [18], [19] | High accuracy due to refined proposal & classification steps | Slower inference |
| | Anchor-based One-stage Detectors | YOLOv3, YOLOv4, YOLOv5 | [20], [21], [22] | Faster than two-stage, good balance of speed/accuracy | High computational cost from anchor boxes |
| | Anchor-free One-stage Detectors | YOLOv8 | [3], [23], [24] | No anchor box overhead, fast and efficient | Slightly lower accuracy in some cases |
| Pruning | Unstructured Pruning (Weight-based) | Weight thresholding | [4], [25], [26] | High compression rate, preserves accuracy | Irregular structure, requires specialized hardware |
| | Structured Pruning (Channel/Layer/Filter) | Network Slimming, ThiNet, Channel Exploration (CHEX), YOLOv5-based Channel Pruning | [27], [28], [29], [30], [31], [32], [33], [34], [35] | Maintains regular topology, suitable for hardware deployment | Lower pruning ratio, huge accuracy loss |
| | Pattern-based Pruning | Yolobile | [36] | Structured weight removal, hardware-accelerator friendly | Dependent on specific compilers/hardware |
| Knowledge Distillation | Logit-based KD (Soft targets) | Hinton's KD, Student-teacher with SoftMax temperature | [37], [38] | Simple implementation, effective for classification | Less guidance for early/mid layers |

## 2.1 Object detection networks

Current CNN-based object detection methods can be categorized into three main types: two-stage detectors, anchor-based one-stage detectors, and anchor-free one-stage detectors. Two-stage detectors, such as [16], [17], [18], [19], operate in two steps: first, they use a region proposal network (RPN) to generate potential object regions, which are then refined and classified in the second stage. While these detectors typically offer higher accuracy than one-stage detectors, they come at the cost of longer inference times. Anchor-based one-stage detectors [20], [21], [22] predict object categories and bounding boxes directly from feature maps, making them more efficient than two-stage detectors. However, they rely on a large number of predefined anchor boxes as reference points, leading to additional computational overhead. To mitigate this, anchor-free one-stage detectors [3], [23], [24] predict the key points and positions of objects without anchor boxes, but at the risk of slightly reduced accuracy.

In the context of aerial imagery, object detection faces additional challenges due to varying object scales, complex backgrounds, and environmental factors such as weather and lighting. Aerial images often contain objects in diverse orientations and sizes, complicating accurate localization. Moreover, the expansive nature of aerial imagery increases the difficulty of detecting densely packed instances and small objects [39]. These complexities necessitate robust and efficient object detection models capable of addressing these challenges effectively. Considering these factors, YOLOv8 [3] emerges as an ideal model for aerial object detection tasks. As an anchor-free one-stage detector which combines speed and accuracy, making it particularly suitable for real-time applications in aerial imagery.

**YOLOv8 Architecture Overview:** YOLOv8, a state-of-the-art object detection algorithm introduced by Ultralytics, builds upon the foundational principles of the YOLO series. Its architecture is structured into four primary components: input, backbone, neck, and head. As an evolution of YOLOv5, YOLOv8 integrates several significant enhancements designed to improve detection performance and address the limitations of its predecessor.

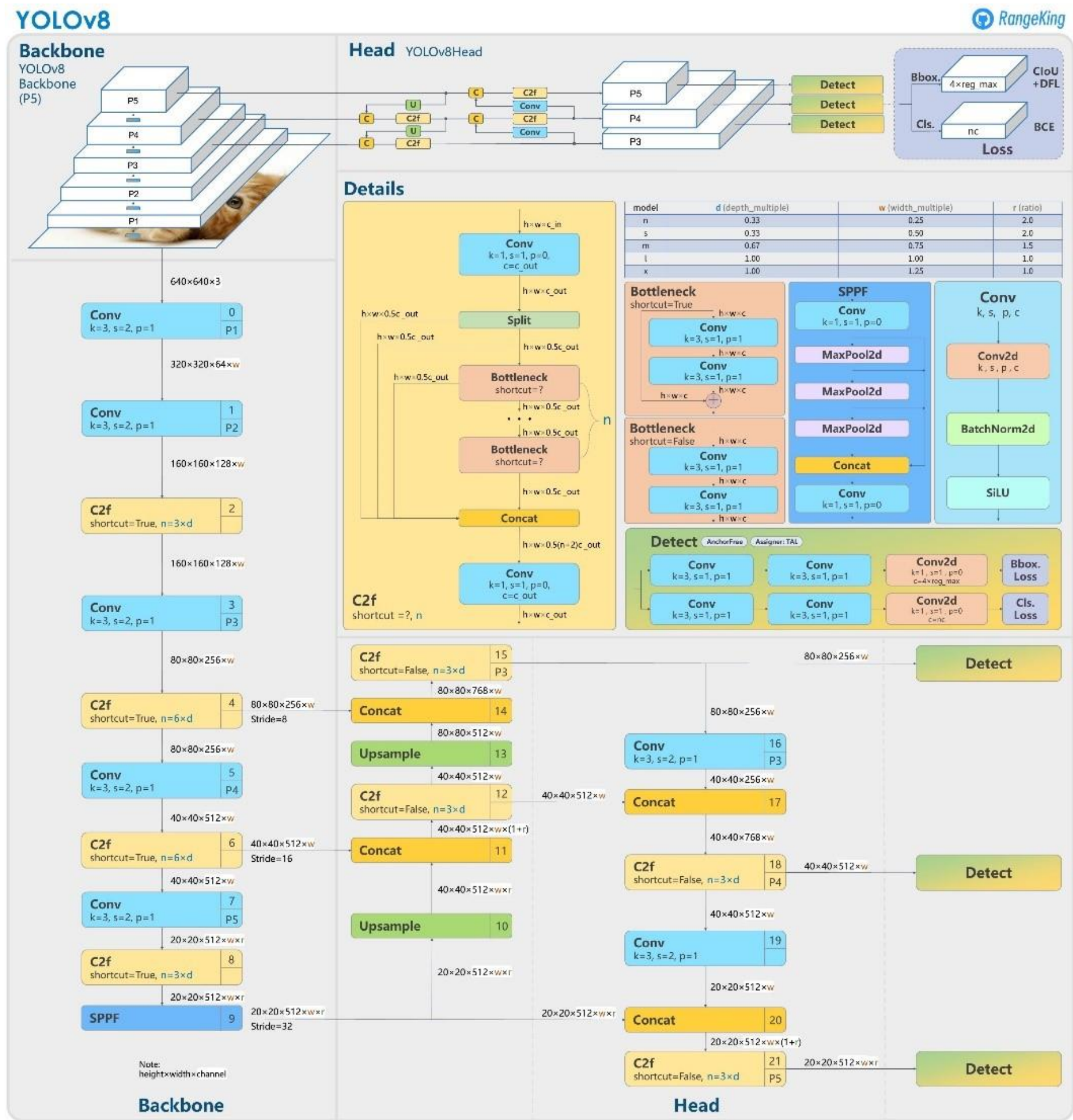

**Fig. 2.** Overview of YOLOv8 architecture showing its components, including the backbone, neck, and head. from [40].

- Backbone and Neck: YOLOv8 introduces a novel network module, C2f, which replaces the C3 module used in YOLOv5. The C2f module strengthens network connectivity and facilitates enriched gradient flow by incorporating additional cross-layer links. These enhancements enable more effective feature extraction and contribute to improved detection accuracy. However, these benefits come with increased model complexity and higher computational requirements, particularly due to the branching structure of the C2f module. The overall architecture, including the C2f module and other components, is depicted in **Fig. 2**.
- Head: YOLOv8 adopts an anchor-free design with a decoupled head, moving away from the traditional anchor-based architecture. The Obj branch present in YOLOv5 is removed, resolving logical inconsistencies between classification and quality as-sessment. This streamlined design results in a more coherent and efficient architecture.

- Loss Function: To enhance optimization, YOLOv8 employs a composite loss function comprising classification BCE, regression CIoU, and distribution focal loss (DFL). This combination enables more robust and precise training.
- Label Assignment: In contrast to the static matching strategy used in YOLOv5, YOLOv8 utilizes a dynamic task-aligned assigner. This approach addresses challenges in sample matching during training, resulting in more effective learning and improved detection outcomes.

While these advancements significantly improve detection performance, they also introduce additional computational demands. To address this, our methodology incorporates pruning strategies targeting convolutional layers, with particular focus on the C2f modules. By reducing the computational burden of the C2f architecture and optimizing other layers, we achieve a lightweight, efficient model with minimal loss in accuracy.

### 2.2 Network Pruning

Network pruning aims to reduce redundancy in the structure and parameters of neural networks, enhancing their efficiency and computational needs. Research in this area focuses on two primary aspects: identifying which components of the network can be pruned and determining how to assess the importance of these components. From the perspective of the components being pruned. Current pruning methods can be classified into unstructured, structured and pattern-based pruning as shown in **Fig. 3**.

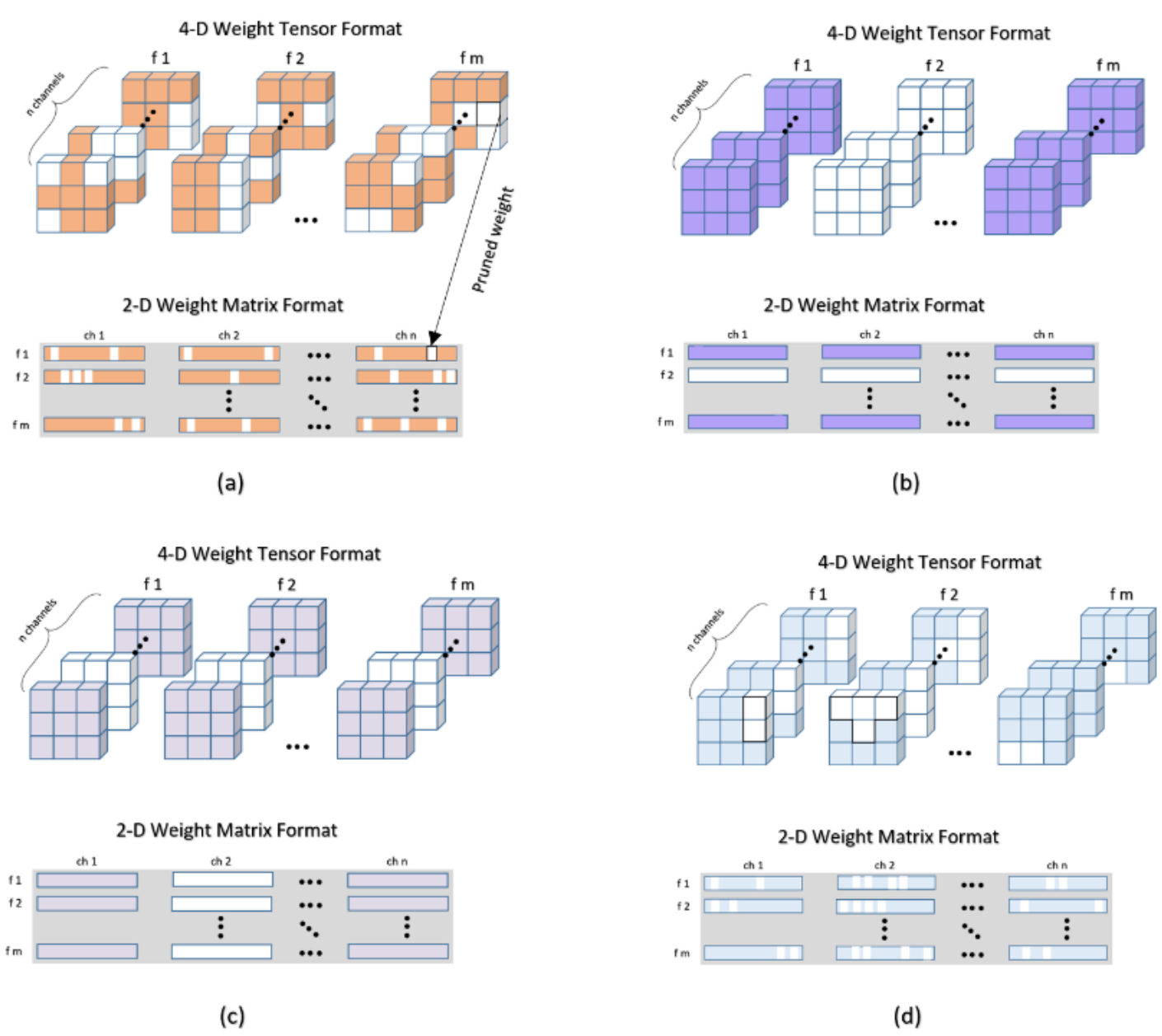

**Fig. 3.** Different pruning methods. a) Unstructured pruning b) Structured layer pruning c) Structured channel pruning d) Pattern-based pruning

Unstructured pruning results in an irregular and unstructured network topology after the pruning process, typically targeting the connection weights between neurons. Unstructured pruning can effectively reduce the number of weights in a model while keeping the loss to a minimum, thereby ensuring that the overall performance remains intact [4]. Additionally [25] and [26] used the absolute value of the weight as the criterion for pruning. The advantage of unstructured pruning lies in its ability to achieve high pruning rates without significantly impacting network accuracy. However, it often requires specialized hardware support, which can increase deployment costs. In contrast, structured pruning maintains the original network topology post-pruning, usually focusing on filters [29], channels [30], or layers [31]. For example [32] employs a unique channel exploration strategy that involves repeatedly pruning and re-growing channels throughout the training process. This dynamic approach mitigates the risk of prematurely removing important channels, allowing for a more flexible and efficient model compression. By utilizing a column subset selection (CSS) formulation for intra-layer pruning and dynamically reallocating channels across layers. [33] Incorporates comprehensive pruning techniques for YOLOv3, Including layer-level and channel-wise pruning to reduce model parameters effectively. In [34], the authors optimize YOLOv4-tiny [35] by pruning 50% of the filters in the final convolutional layers, reducing memory and computation requirements. In [27] the authors propose a channel pruning strategy for YOLOv5 [22] network to detect safety helmets and anti-slip shoes efficiently. The method involves three steps: sparse training with L1 regularization to identify low-importance parameters, channel pruning to remove less significant channels and reduce network width, and fine-tuning to recover any precision loss post-pruning. Following the strategy outlined in [27], [28] develops an enhanced version by combination of pruning strategies and data preprocessing to improve YOLOv5's performance. They also compare normal and regular pruning (in this method the number of filters after pruning is a multiple of 8 for better hardware deployment) at varying rates, finding that while normal pruning achieves higher mAP values, regular pruning is more suitable for hardware constraints. To enhance performance, five data preprocessing methods were tested, with random shape training showing the most significant accuracy improvement. Ultimately, the authors select random shape, random angle, and saturation variation for balanced accuracy and efficiency in object detection tasks. The primary advantage of structured pruning is its independence from specialized hardware; however, it generally achieves lower pruning rates compared to unstructured methods. Pattern-based pruning focuses on removing weights in structured patterns, such as blocks or groups, from convolutional filters, improving both efficiency and compatibility with hardware accelerators. Unlike unstructured pruning, which creates irregular sparsity, pattern-based pruning ensures regular weight removal, enabling better hardware optimization. For instance, [36] introduce a compression-compilation co-design approach in YOLOv5 called Yolobile, which applies structured pruning patterns to optimize real-time object detection on mobile devices. This method combines pattern pruning with compiler-level optimizations, reducing computational demands while maintaining accuracy. Although this strategy achieves faster inference and better hardware utilization, it relies on specific hardware and compiler support, which can limit its applicability in diverse environments.

Inspired by the methods presented in [27] and [28], this study adopts a flexible three-step approach to enhance the efficiency of the YOLOv8 network. The method begins with sparse training using L1 regularization to identify low-importance parameters, followed by channel pruning to remove less significant channels and reduce the network width. Instead of relying heavily on fine-tuning, knowledge distillation is employed as a key step to recover accuracy by transferring knowledge from a high-performance teacher network to the compressed student model. This structured pruning and knowledge distillation strategy ensures a flexible and efficient framework, compatible with diverse hardware setups and free from the need for specific accelerators.

### 2.3 Knowledge Distillation

Knowledge distillation (KD) is a technique designed to transfer knowledge from a large, high-performing teacher network to a smaller student network, improving the performance of the compact model while maintaining efficiency. Initially introduced by Hinton et al. [41] for classification tasks, KD has since gained widespread application across domains such as computer vision [10], [42], [43], natural language processing [44], [45], and speech recognition [46]. Research in this area primarily focuses on two key aspects: identifying which components of the network should be considered as knowledge and determining how to assess whether the student network has effectively learned this knowledge, which is often reflected in the design of the distillation loss function. Current distillation methods can be categorized based on what is selected as knowledge: 1) using the final class output of the teacher network, as seen in [37], [38]) utilizing intermediate feature layers from the teacher network, as demonstrated in [47], [48]; and 3) leveraging the structural relationships between layers of the teacher network, as discussed in [49].

While KD enhances the performance of smaller models, it does not inherently reduce model size or computational demands as pruning does. In recent years, researchers have explored hybrid approaches that combine KD with other model compression techniques, such as pruning and quantization. The combination of KD with quantization has been particularly effective in certain applications [50], [51], [52], leveraging KD to recover performance losses introduced by reduced precision. However, the integration of KD and pruning remains less explored, particularly in the context of advanced anchor-free single-stage object detection networks like YOLOv8. This represents a promising area for research, as pruning effectively reduces model size and computational requirements, while KD helps mitigate the accuracy loss often associated with aggressive pruning.

This study addresses this gap by incorporating structured pruning alongside channel-wise KD tailored for YOLOv8. Inspired by [53], we extract knowledge from intermediate feature layers of the teacher network and transfer it to the student model. This approach enables the pruned model to retain high accuracy while remaining computationally efficient. By focusing on channel-wise KD, the student model learns to prioritize the most critical features, achieving improved performance without requiring specialized hardware or accelerators. This makes the method particularly suitable for resource-constrained environments, such as aerial imagery object detection, where computational resources are often limited.

# 3 Methodology

We propose a three-stage compression pipeline for YOLOv8 to reduce the model's computational overhead while preserving its high detection accuracy, particularly for aerial imagery tasks. The pipeline consists of:

1. Dynamic Sparse Training: Identifies and ranks less critical channels by enforcing sparsity in the network.
2. Structured Pruning: Removes redundant channels based on importance rankings, improving computational efficiency.
3. Channel-Wise Knowledge Distillation (CWD): Refines the compressed model by transferring knowledge from the original model to ensure minimal performance degradation.

The rationale for each component and their integration into the YOLOv8 architecture are detailed in the following sections.

### 3.1 Sparse Training with Dynamic Sparsity Scheduling

To prepare YOLOv8 for pruning, sparse training is employed to identify less critical channels in convolutional layers. In the design of YOLOv8 architecture, each convolutional structure consists of three network layers: the convolutional layer, the Batch Normalization layer, and the activation layer. This process focuses on the BN (BatchNorm2d) layers, where the scaling factors ($\gamma$) serve as indicators of channel importance. The BN layer normalizes the input activations $\hat{z}_i$ as follows:

$$\hat{z}_i = \frac{z_i - \mu_B}{\sqrt{\sigma^2{}_B + \epsilon}} \quad (1)$$

Where $z_i$ represents the activations, $\mu_B$ and $\sigma_B$ are the mean and variance of the batch respectively, and $\epsilon$ is a small constant to ensure numerical stability. After normalization, a linear transformation is applied using the learnable scaling factor $\gamma$ and the shifting parameter $\beta$, producing the output as:

$$\hat{z}_{out} = \gamma \hat{z}_i + \beta \quad (2)$$

The scaling factor $\gamma$ determines the relative importance of each channel, where smaller values of $\gamma$ correspond to channels with less significant activations. Sparse training introduces L1 regularization to $\gamma$ to enforce sparsity, encouraging small values of $\gamma$, which can be pruned without significantly impacting the network's performance. The objective is to identify channels that contribute less to the final output, allowing for their removal during the pruning phase. This process effectively ranks the channels based on their importance, leading to a more efficient and compact model while preserving essential features for accurate predictions.

A gradual sparsity enforcement strategy is adopted to ensure stable training and prevent abrupt drops in detection performance. The sparsity rate is incrementally increased during training using the following formula where $e$ is Current training epoch.

$$\text{Sparsity Rate (e)} = \text{Initial Rate} \times (1 - 0.9.\frac{\text{e}}{\text{Total Epochs}}) \quad (3)$$

This smooth transition facilitates the network's adaptation to sparsity, preserving its detection capabilities while preparing it for pruning.

### 3.2 Structured Pruning

Structured pruning is applied after sparse training to reduce the computational complexity of the model. This step focuses on convolutional layers, which are the most computationally expensive components. By selectively removing channels, the model’s size is reduced without significantly impacting detection performance.

The importance of each channel is determined by the magnitude of its Batch Normalization scaling factor ($\gamma$), Channels with smaller importance scores are considered less critical and candidate of pruned.

A pruning ratio is applied uniformly across all convolution layers to ensure a balanced trade-off between compression and accuracy. In this work, a pruning ratio of 50% is used, where 50% of channels with the smallest importance values are pruned in each layer. This fixed pruning ratio was chosen based on the need for a significant reduction in computational load while maintaining essential feature representations.

Each convolution layer is pruned individually, ensuring that important features in key network components (e.g., C2f blocks and fusion stages like Upsample + Concat) are preserved.

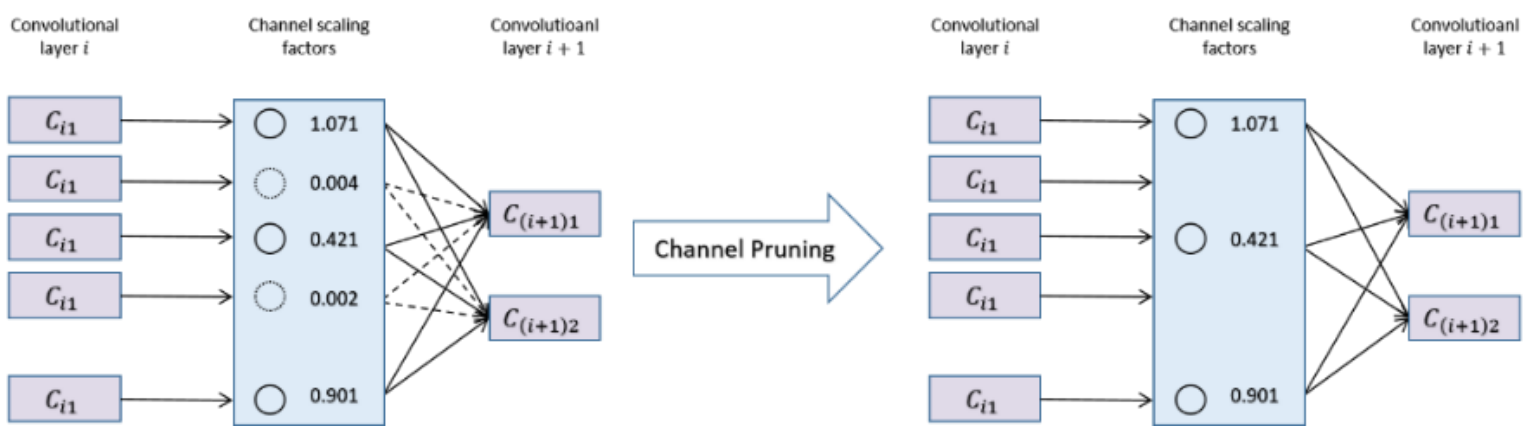

**Fig. 4.** Schematic representation of the channel pruning algorithm

This structured pruning process as shown in **Fig. 4**, results in a model that is both smaller and more efficient, while still retaining critical features for accurate predictions. The next phase involves fine-tuning the pruned model and then use channel-wise knowledge distillation to recover any performance degradation.

### 3.3 Channel-Wise Knowledge Distillation (CWD)

In this study, we adopt the concept of channel-wise knowledge distillation (CWD) as introduced in [53] to recover performance loss caused by pruning, particularly focusing on the C2F modules in the neck of the YOLOv8 architecture. Pruning often disproportionately impacts critical channels responsible for dense predictions, such as those within the C2F neck module, which plays a vital role in object detection by aggregating and refining features. To address this, we apply the channel-wise distillation strategy to align the activations of corresponding channels between a stronger teacher network and the pruned student network as shown in **Fig. 5**.

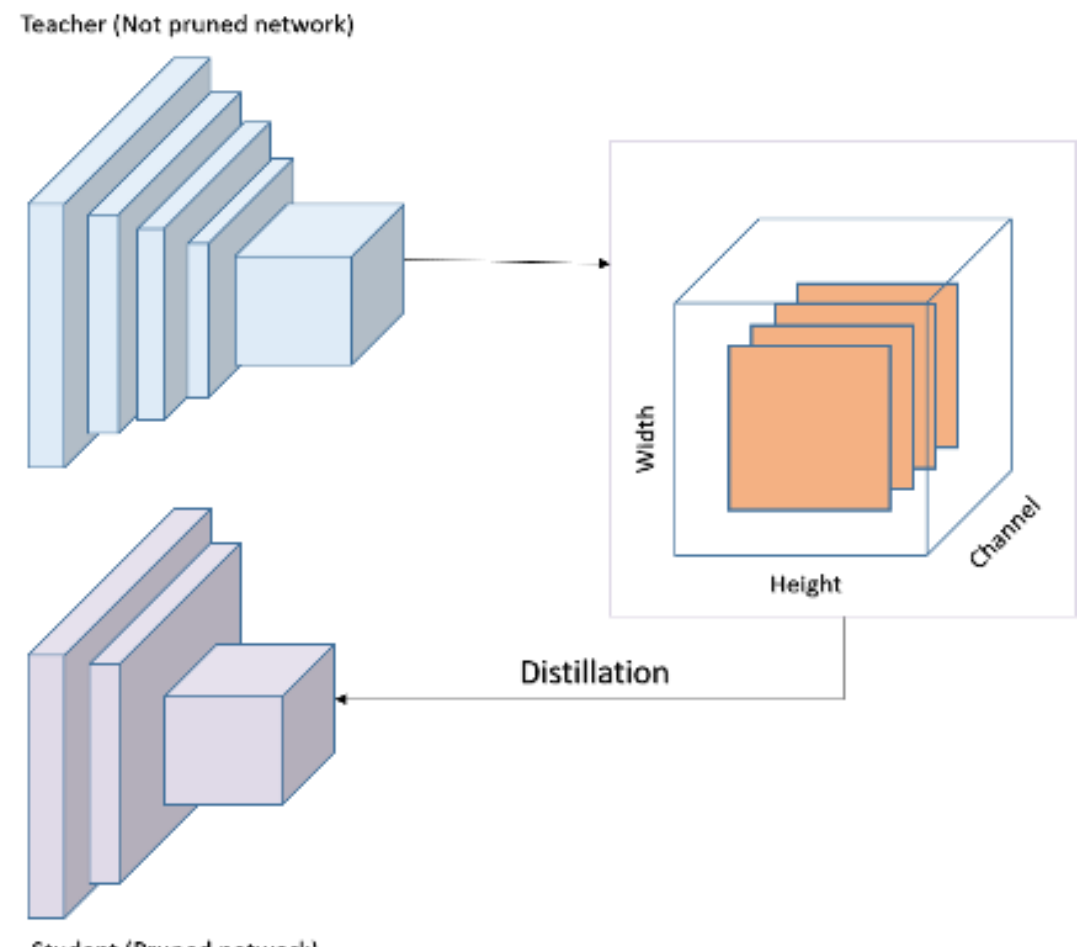

**Fig. 5.** Channel distribution distillation aligns each channel of student's feature maps to that of the teacher network by minimizing the KL divergence

Following the established CWD paradigm, we convert the activation maps of each channel into probability distributions across their spatial dimensions. This enables the use of the Kullback-Leibler (KL) divergence as the distance metric to measure discrepancies between the teacher and student activations. For a given channel $c$, the probability distribution is calculated as:

$$\varphi(y_c)_i = \frac{\exp({y_i}^c/\tau)}{\sum_{j=1}^{W.H}\exp({y_j}^c/\tau)} \tag{4}$$

Where $\tau$ is a temperature parameter that controls the softness of the distribution, $i$ indexes the spatial locations, and $W$ and $H$ denote the width and height of the activation map. A larger $\tau$ broadens the spatial focus, resulting in a softer distribution.

Then the channel-wise distillation loss is defined as:

$$L_{CWD} = \frac{\tau^2}{C}\sum_{c=1}^{C}\sum_{i=1}^{W.H}\varphi\left({y^c}_T\right).\log\left(\frac{\varphi({y^c}_T)_i}{\varphi({y^c}_S)_i}\right) \tag{5}$$

Where ${y^c}_T$ and ${y^c}_S$ represent the activation maps of channel $c$ from the teacher and student networks, respectively. The asymmetric nature of the KL divergence ensures that high-activation regions (foreground) are emphasized during training, while low-activation regions (background) have reduced influence. This objective encourages the student to replicate the teacher's strongest activations, ensuring that critical features are preserved in the pruned model. This property is particularly beneficial for dense prediction tasks like object detection.

In our approach, we primarily target the C2F modules within the neck of YOLOv8 for channel-wise distillation, as these layers (e.g., layers 12, 15, 18, and 21) are criti-cal for aggregating and refining multi-scale features. By recovering these pruned channels, we aim to enhance the detection capabilities of the student network. We evaluate several configurations for the distillation process:

- Distilling from Backbone and Neck C2F Modules: Starting with the primary layers in the backbone and progressively including C2F modules in the neck.
- Distilling from C2F Neck Modules Only: Focusing exclusively on the neck layers, which yielded the best results in our experiments.
- Teacher-Student Pairing: Employing a stronger teacher network (e.g., YOLOv8-L) to guide a smaller student network (e.g., YOLOv8-M).

Our results indicate that channel-wise distillation from the neck's C2F modules is particularly effective. These layers are pivotal for synthesizing high-level semantic features, and recovering their functionality significantly boosts the performance of the pruned model. The targeted recovery not only compensates for the pruning-induced loss but also ensures that the student network maintains competitive detection accuracy with reduced computational complexity.

By leveraging the asymmetric properties of KL divergence and focusing on critical layers in the C2F Neck, our approach demonstrates the effectiveness of channel-wise knowledge distillation in addressing performance degradation caused by pruning. This method highlights the importance of selectively targeting essential network components to achieve a balance between efficiency and accuracy in anchor-free single-stage object detection models.

## 4 Results

### 4.1 Experimental environment and parameter set

The experimental framework for all model training and evaluation was conducted in PyTorch on NVIDIA GPUs, including the RTX 3090, RTX 4090, and Quadro RTX 8000. These GPUs were used at different stages of experimentation based on availability, with each model being trained on a single GPU at a time. Inference speed was measured on an NVIDIA GeForce GTX 1080 to simulate deployment scenarios on edge devices with limited processing capabilities.

Our approach was trained and tested using the hyperparameters listed in **Table 2** For sparse training, the initial sparsity rate was set to 0.005, and a structured pruning ratio of 50% was applied afterward. Unless otherwise specified, other training and testing configurations followed YOLOv8 default settings. To enhance robustness and generalization, the built-in data augmentation techniques of YOLOv8 were utilized. These augmentations help the model adapt to varied object appearances and improve detection under diverse conditions.

**Table 2.** Network training hyperparameters

| Train parameters | Configuration |
|---|---|
| Learning rate | 0.01 |
| Momentum | 0.937 |
| Weight decay | 0.0005 |
| Batch size | 8 |
| Image size | 1024x1024 |
| Epochs | 100 |

### 4.2 Dataset

For training and evaluating the model, we used the VisDrone dataset, which offers a diverse set of aerial images captured in both urban and rural settings. It includes various object categories such as vehicles, pedestrians, and bicycles, often appearing in crowded scenes with many small objects. These challenges make VisDrone a strong benchmark for testing detection performance in real-world conditions.

We followed the official dataset split, using 6,471 images for training, 548 for validation, and 1,610 for testing. **Table 3** shows the distribution of object instances based on their area, highlighting the wide range of object sizes in the dataset.

**Table 3.** The distribution of data based on area in VisDrone dataset

| Size | $<32^2$ | $32^2 \sim 96^2$ | $> 96^2$ | $<200^2$ | $200^2 \sim 400^2$ | $400^2 <$ |
|---|---|---|---|---|---|---|
| #Samples | 306262 | 159999 | 23703 | 487887 | 2035 | 42 |

### 4.3 Experimental results and analysis

In this section, we present and analyze the performance of the proposed YOLOv8 compressing approach and its variants across multiple metrics, including AP, AP50, AP75, $AP_{small}$, $AP_{medium}$, $AP_{large}$, model size, number of parameters, GFLOPs, and MACs. These metrics provide a holistic assessment of detection accuracy, computational efficiency, and adaptability. Results are analyzed based on the VisDrone dataset and compared with state-of-the-art methods.

**Sparsity-Aware Training and Pruning Results.**

To assess the effectiveness of sparsity-aware training, we evaluated its impact on various YOLOv8 model variants (n, s, m, l, x) using the VisDrone dataset. This approach, which integrates structured sparsity during training, led to consistent improvements across most performance metrics. As shown in **Table 4** sparsity-aware training enhanced the average precision (AP) for nearly all model scales. For example, the AP of YOLOv8-n increased from 22.1 to 22.4, and YOLOv8-l improved from 29.9 to 30.2.

Notably, the gains were more pronounced in detecting small and medium-sized objects—an essential factor in aerial image analysis where dense object distributions are common. Detailed results for AP50 are provided in **Table 5** to optimize YOLOv8 for aerial object detection, we applied channel pruning to remove less significant chan-nels based on the scaling factors from batch normalization layers. After pruning, a finetuning phase was performed for 100 epochs to allow the network to adapt to the reduced structure and recover potential performance losses. All the reported results for the pruned models reflect the performance after this fine-tuning step. For our experiments, we primarily focus on the YOLOv8-m variant, as it offers the best trade-off between computational efficiency and detection accuracy. Compared to the smaller (n, s) and larger (l, x) variants, YOLOv8-m provides a balanced performance profile, making it an ideal candidate for optimization under real-world constraints such as edge device deployment.

**Table 4.** Impact of sparsity-aware training on AP (VisDrone dataset)

| Model | AP (Baseline) | AP (Sparse) | Δ AP |
|---|---|---|---|
| YOLOv8-n | 22.1 | **22.4** | +0.3 |
| YOLOv8-s | 26 | **26.3** | +0.3 |
| YOLOv8-m | 28.3 | **28.5** | +0.2 |
| YOLOv8-l | 29.9 | **30.2** | +0.3 |
| YOLOv8-x | **30.6** | 30.5 | -0.1 |

**Table 5.** Impact of sparsity-aware training on AP50 (VisDrone dataset)

| Model | AP50 (Baseline) | AP50 (Sparse) | Δ AP50 |
|---|---|---|---|
| YOLOv8-n | 40 | **40.4** | +0.4 |
| YOLOv8-s | 46.5 | **46.9** | +0.4 |
| YOLOv8-m | 50.2 | **50.6** | +0.4 |
| YOLOv8-l | 52.7 | **53** | +0.3 |
| YOLOv8-x | 53.4 | **53.5** | +0.1 |

As shown in **Table 6** channel pruning effectively reduced the computational complexity and model size while maintaining acceptable detection performance. Specifically, after applying 50% pruning, the AP of YOLOv8-m decreased slightly from 28.5 to 26.6. Similar trends were observed across most metrics, including $AP_{small}$ (dropping from 41.3 to 37.8) and $AP_{medium}$ (from 66.2 to 62.1).

Interestingly, despite a general decrease in performance on smaller and medium objects, the $AP_{large}$ metric improved significantly from 63.0 to 69.7. This suggests that pruning may have indirectly enhanced the model's ability to detect larger objects, possibly by reallocating capacity toward dominant features. This trade-off between computational efficiency and detection accuracy highlights the importance of carefully balancing model compression strategies, particularly when targeting deployment on resource-constrained devices.

Additionally, we evaluated the impact of pruning on model complexity in terms of the number of parameters, MACs, FLOPs, and model size, summarized in **Table 7** After pruning, the number of parameters dropped from 25.86 million to 6.85 million. Correspondingly, the MACs and FLOPs reduced from 101.00 G to 34.51 G, and from 49.6 billion to 13.3 billion, respectively. The model size also decreased substantially, from 49.6 MB to 13.3 MB. These results clearly demonstrate the effectiveness of pruning in achieving a lighter and faster model with minimal loss in detection capability, making it highly suitable for aerial object detection tasks in real-world scenarios.

**Table 6.** Detection performance of YOLOv8-m before and after 50% channel pruning on the VisDrone dataset.

| Model | AP | AP50 | AP75 | APsmall | APmedium | APlarge |
|---|---|---|---|---|---|---|
| YOLOv8-m (Base) | 28.3 | 50.2 | 27.8 | 41.3 | 66.2 | 63 |
| YOLOv8-m (Pruned) | 26.6 | 47.3 | 25.9 | 37.8 | 62.3 | 69.7 |
| Δ | -1.9 | -2.9 | -1.9 | -3.5 | -3.9 | +6.7 |

**Table 7.** Computational and efficiency metrics of YOLOv8-m before and after 50% channel Pruning.

| Model | #parameters | MAC | FLOPs | Model size | FPS |
|---|---|---|---|---|---|
| YOLOv8-m (Base) | 25,862,110 | 101.0024 | 49.6 | 49.6 | 17 |
| YOLOv8-m (Pruned) | 6,845,710 | 34.51155 | 13.3 | 13.3 | 44 |
| Δ | -19,016,400 | -66.49 | -36.3 | -36.3 | +27 |

Pruning resulted in a substantial reduction of approximately 73% in the number of parameters, 66% in MAC, 73% in FLOPs, and 73% in model size. Additionally, FPS improved by %%, indicating better inference efficiency. However, there was a slight decrease in detection performance, with a slight reduction in detection performance. Despite these performance trade-offs, the pruned model demonstrates a more efficient use of resources, making it better suited for deployment on resource-constrained devices.

**Knowledge Distillation Results**

To recover the performance lost during pruning, Channel-Wise Knowledge Distillation (CWD) was applied in several configurations. The first configuration (C1) applies CWD to the neck layers, which are typically considered key for feature aggregation. The second configuration (C2) extends this approach by incorporating CWD from all critical convolution layers (C2f), aiming to leverage more comprehensive knowledge transfer. The third configuration (C3) explores distillation from a stronger teacher model, YOLOv8-l, to guide the pruned YOLOv8-m model in improving accuracy.

These configurations are designed to evaluate the impact of knowledge transfer at different levels of the network, with each approach intended to address specific performance aspects. The results presented in **Table 8** demonstrate that CWD can effectively recover some of the performance lost during pruning, particularly in the AP50 and $AP_{large}$ metrics. Among the configurations, C1 achieved the most notable improvement, with an AP of 26.8, compared to C2 (26.1) and C3 (26.6), which showed slightly lower performance. These findings underscore the importance of selecting the right layers for knowledge transfer during distillation, suggesting that targeting specific network components can lead to better results.

**Table 8.** Comparison of detection performance across different KD configurations

| Model | AP | AP50 | AP75 | $AP_{small}$ | $AP_{medium}$ | $AP_{large}$ |
|---|---|---|---|---|---|---|
| Baseline | 28.3 | 50.2 | 27.8 | 41.3 | 66.2 | 63 |
| Pruned | 26.6 | 47.3 | 25.9 | 37.8 | 62.1 | **69.7** |
| CWD-S1 | **26.8** | **47.9** | **26** | **38.1** | **63.5** | 64.7 |
| CWD-S2 | 26.1 | 46.8 | 25.4 | 37.6 | 62.1 | 57.8 |
| CWD-S3 | 26.6 | 47.4 | 25.9 | 38 | 62.9 | 63.9 |

*Effect of Temperature (T) Parameter on Network Accuracy:* The temperature parameter (T) in knowledge distillation critically influences the smoothness of the teacher network's output distribution. To investigate its effect, a range of temperature values (T = 1 to 10) was explored on the YOLOv8-m model. As shown in **Fig. 6**, moderate temperature values generally led to slight improvements in AP and AP50, while excessively high or low values tended to cause fluctuations.

The optimal temperature was observed at T = 6, where both AP and AP50 achieved peak performance, suggesting an improved balance between knowledge transfer smoothness and model generalization. Lower values, such as T = 1 and T = 2, provided comparable but slightly inferior results, while temperatures beyond T = 6 led to marginal declines in detection accuracy.

These findings indicate that careful tuning of the temperature parameter is necessary to maximize the benefits of distillation, particularly for maintaining high performance on dense object detection tasks.

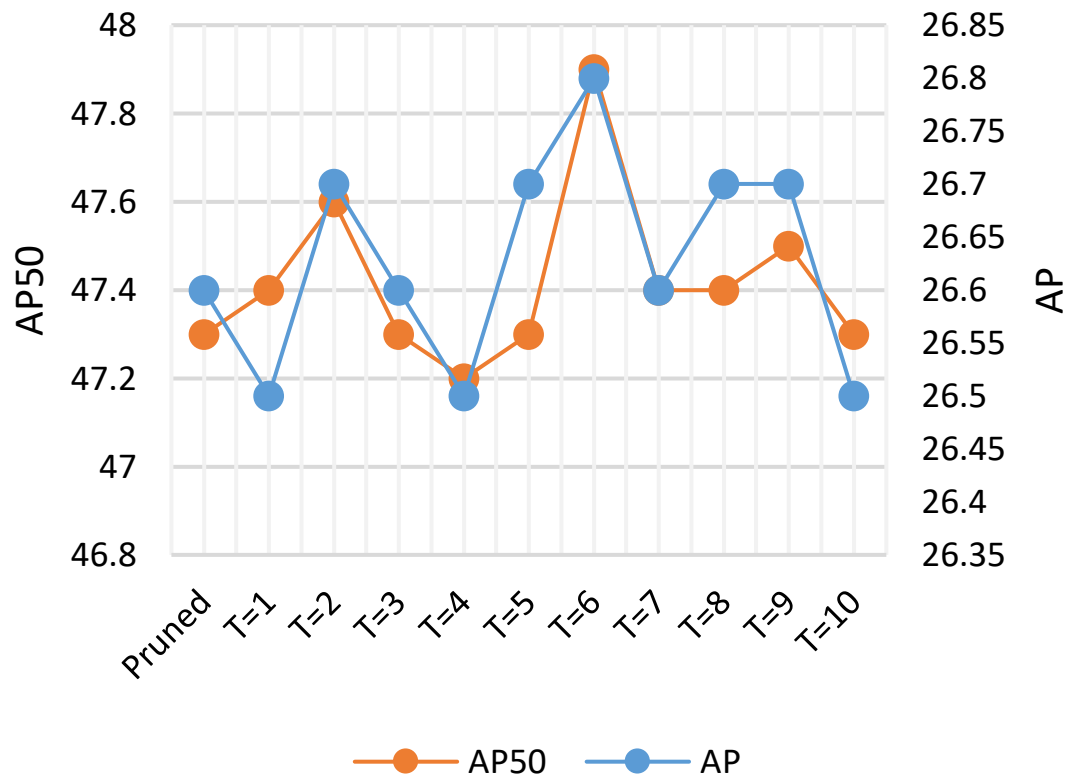

**Fig. 6.** Effect of temperature parameter on AP and AP50

*Effect of α Parameter on Network Accuracy*: The α parameter, which weights the CWD loss, was tuned to enhance the pruned YOLOv8-m model. As shown in **Fig. 7** setting α = 0.5 yielded the best results, improving AP from 26.6 to 26.8 and AP50 from 47.3 to 47.9. In contrast, lower (α = 0.3) and higher (α = 0.8) values slightly degraded performance. These findings demonstrate the importance of properly balancing the distillation loss during training.

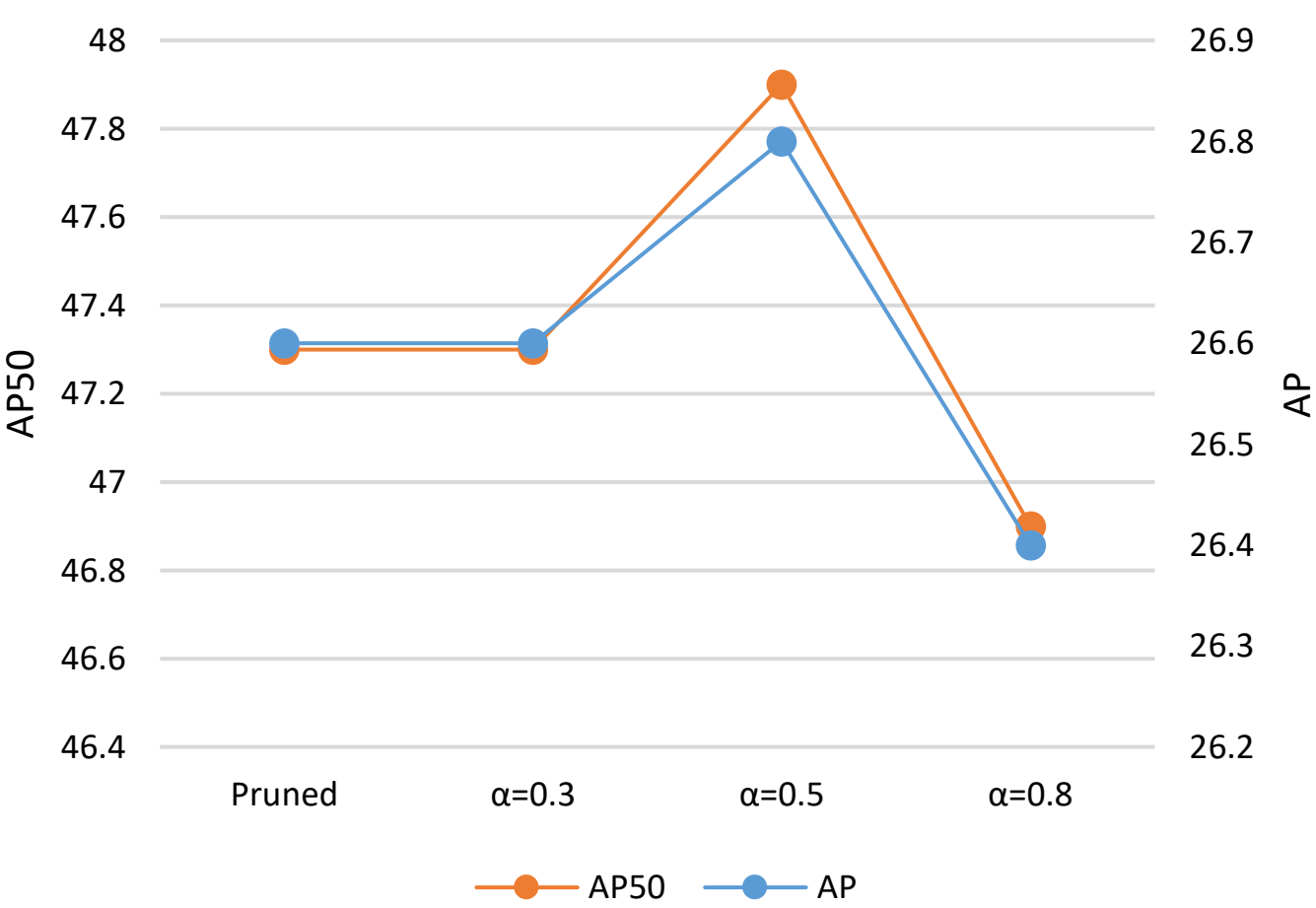

**Fig. 7.** Effect of α parameter on AP and AP50

*Comparison of YOLOv8 Variants: Model Complexity, Computational Cost, and Detection Accuracy:* The effects of our compression approach on different YOLOv8 variants are summarized in this section. As shown in **Table 9**, pruning significantly reduced model parameters and size, with YOLOv8-m achieving over 73% reduction in both. **Table 10** shows notable reductions in GFLOPs and MACs, improving computational efficiency without heavy sacrifices in performance and finally indicates only a slight drop in AP and AP50 after pruning the model and using channel-wise knowledge distillation (as shown in **Fig. 8** and **Fig. 9** demonstrating that our method effectively balances accuracy and efficiency across the models.

**Table 9.** Model complexity comparison

| Model | #layers | Parameters (Base) | Parameters (Ours) | Reduction (%) | Model size (Base) | Model size (Ours) | Reduction (%) |
|---|---|---|---|---|---|---|---|
| YOLOv8-n | 225 | 3,007,598 | 910,830 | -69.71% | 6 | 2.1 | -65% |
| YOLOv8-s | 225 | 11,129,454 | 3,007,598 | -72.97% | 21.5 | 6 | -72.09% |
| YOLOv8-m | 295 | 25,862,110 | 6,845,710 | -73.51% | 49.6 | 13.3 | -73.18% |
| YOLOv8-l | 365 | 43,614,318 | 19,531,246 | -55.21% | 83.6 | 37.6 | -55.02% |
| YOLOv8-x | 365 | 68,133,198 | 30,280,782 | -55.55% | 130 | 61 | -53.07% |

**Table 10.** Computational cost comparison

| Model | GFLOPs (Base) | GFLOPs (Ours) | Reduction (%) | MAC (Base) | MAC (Ours) | Reduction (%) |
|---|---|---|---|---|---|---|
| YOLOv8-n | 8.1 | 2.1 | -74.07% | 10.43315 | 4.456598 | -57.28% |
| YOLOv8-s | 21.5 | 6 | -72.09% | 36.56131 | 4.456598 | -87.81% |
| YOLOv8-m | 49.6 | 13.3 | -73.18% | 101.0024 | 34.51155 | -65.83% |
| YOLOv8-l | 83.6 | 37.6 | -55.02% | 211.3927 | 68.19593 | -67.73% |
| YOLOv8-x | 130 | 61 | -53.07% | 329.9925 | 106.3743 | -67.76% |

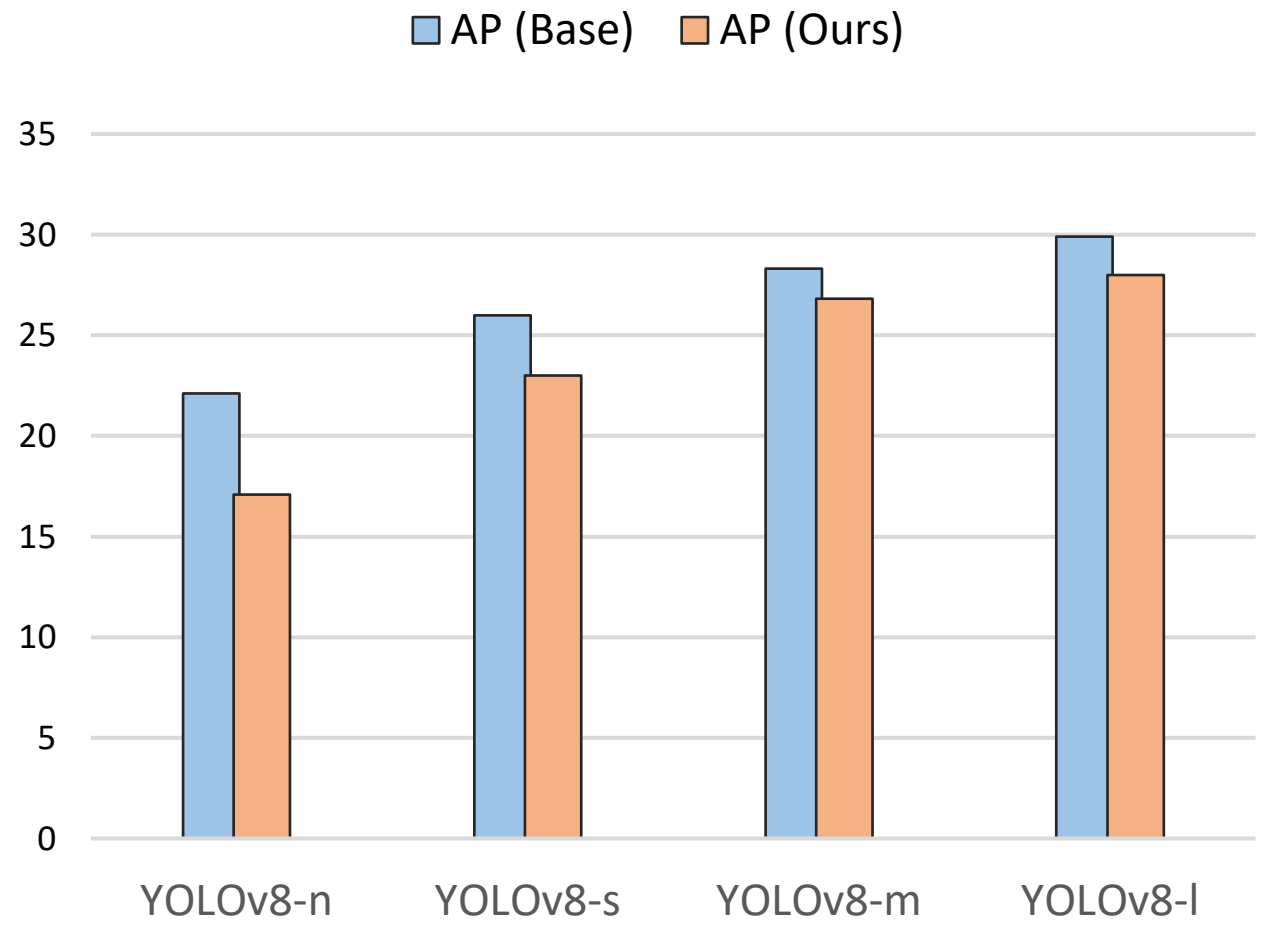

**Fig. 8.** Detection accuracy comparison (AP)

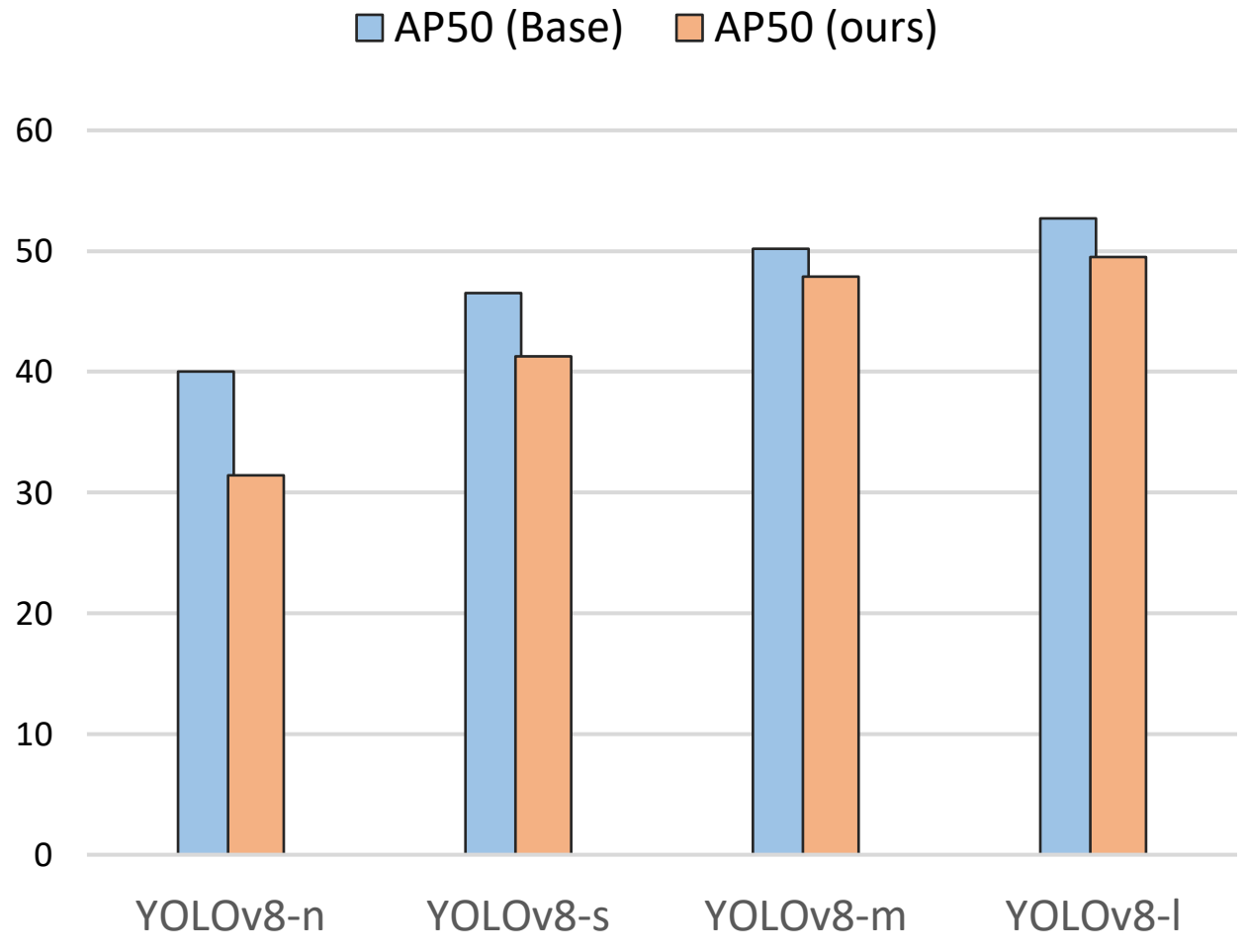

**Fig. 9.** Detection accuracy comparison (AP50)

*Comparison with State-of-the-Art and Recent YOLO Variants:* As shown in **Fig. 10** and **Table 11**, our method demonstrates a superior trade-off between detection accuracy and model complexity when compared against a wide range of SOTA models. In the figure, the orange trend line representing parameter count drops significantly for our method (6.8M) compared to medium-scale competitors like YOLOv9m and YOLOv11m (~20M), while the detection accuracy (represented by the bars) remains competitive or even superior. When compared to the latest iterations of the YOLO family, the advantages of our proposed framework become clear. Standard medium-scale models such as YOLOv9m and YOLOv11m achieve strong accuracy (~47.3%) but do so with a heavy parameter count of approximately 20.0 million. Conversely, our method prunes the YOLOv8 structure down to just 6.8 million parameters—a reduction of roughly 66% (3× smaller)—while achieving a higher AP50 of 47.9%.

Crucially, our advantage extends to specialized drone-specific networks. As detailed in **Table 11**, our method outperforms SL-YOLO [54], a recent SOTA model designed for aerial detection. We achieve higher accuracy (47.9% vs. 46.9%) while utilizing significantly fewer parameters (6.8M vs. 9.6M).

This efficiency allows us to address the critical trade-off between input resolution and speed. While competitors like YOLOv11m and SL-YOLO must operate at standard resolution (640×640) to maintain real-time performance, our proposed method reduces the computational burden enough to allow us to adopt a higher input resolution (1024×1024). This strategy enables the model to capture fine-grained details of small aerial objects—surpassing the accuracy of 640-pixel models—while maintaining a parameter footprint smaller than even the 'Small' versions of modern YOLOs. This positions our approach as a highly effective solution, delivering top-tier accuracy with the efficiency required for real-time edge deployment.

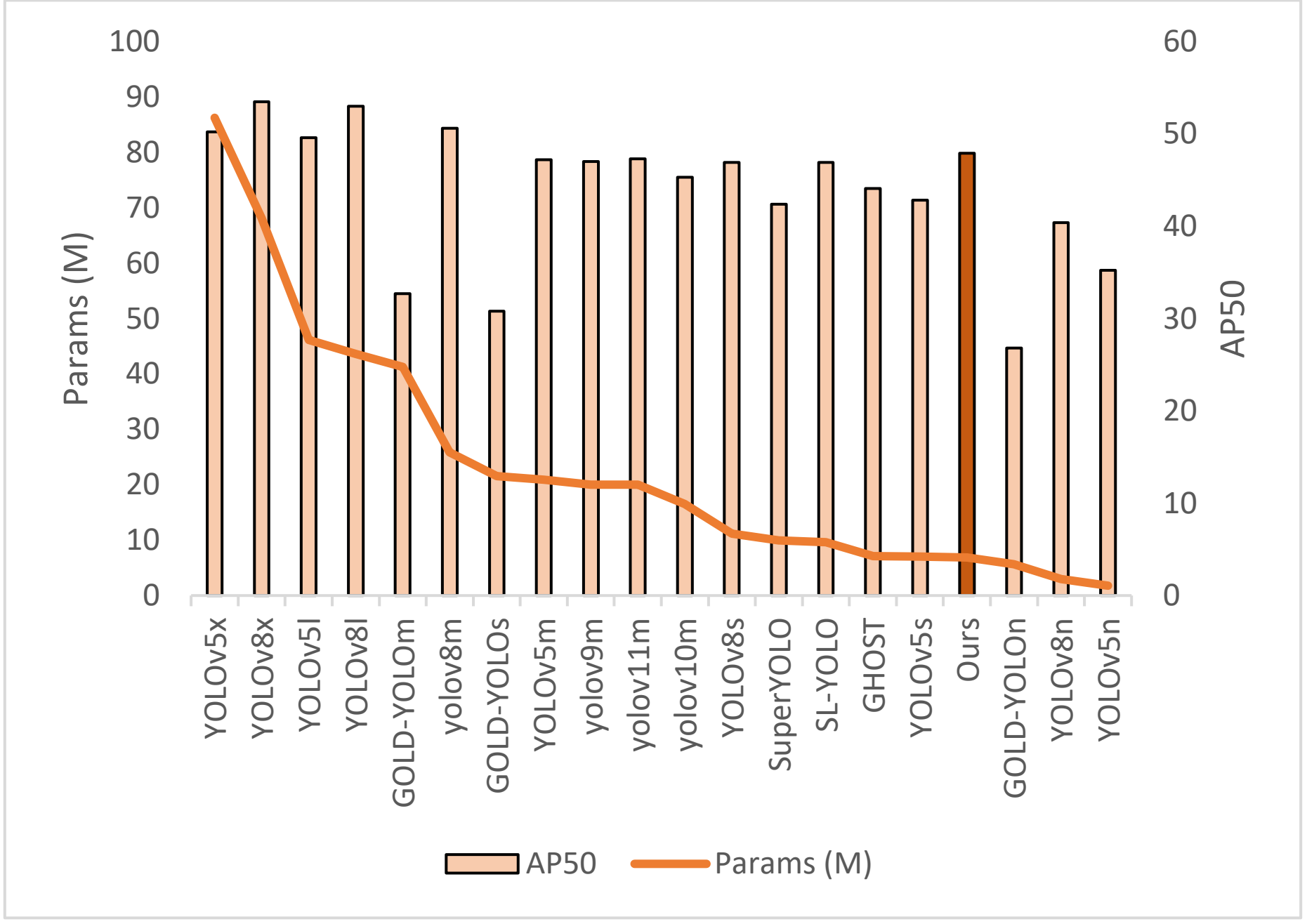

**Fig. 10.** Trade-off between detection accuracy and model size: a comparative analysis of SOTA model variants and our method

**Table 11.** Comparison with State-of-the-Art and recent YOLO variants on the VisDrone2019-val dataset. Results for YOLOv9, v10, v11, and SL-YOLO are cited from [54]

| Model | Input Size | Params (M) | AP50 |
|---|---|---|---|
| Yolov8-m (baseline) | 1024 | 25.8 | 50.2 |
| yolov8-m | 640 | 25.8 | 46.7 |
| Yolov9-m | 640 | 20.0 | 47.0 |
| Yolov10-m | 640 | 16.5 | 45.3 |
| Yolov11-m | 640 | 20.0 | 47.3 |
| SL-YOLO | 640 | 9.6 | 46.9 |
| **Ours** | **1024** | **6.8** | **47.9** |

We further applied TensorRT as an additional optimization step. This lightweight deployment enhancement slightly improved inference speed without affecting detection accuracy, showing that TensorRT can be seamlessly integrated into our pipeline for even better runtime performance on resource-constrained devices **Table 12**.

**Table 12.** Results of basic and compressed networks before and after using TensorRT

| Model | TRT | AP | AP50 | AP75 | $AP_{small}$ | $AP_{medium}$ | $AP_{large}$ | FPS |
|---|---|---|---|---|---|---|---|---|
| Base | × | 28.3 | 50.2 | 27.8 | 41.3 | 66.2 | 63 | 26 |
| Base | ✓ | 28.1 | 50 | 27.8 | 41.1 | 66 | 61.8 | 47 |
| Pruned | × | 26.6 | 47.3 | 25.9 | 37.8 | 62.1 | 69.7 | 44 |
| Pruned | ✓ | 26.6 | 47.4 | 25.8 | 38 | 62.2 | 65.4 | 58 |
| Ours | × | 26.8 | 47.9 | 26 | 38.1 | 63.5 | 64.7 | 45 |
| Ours | ✓ | 26.7 | 47.6 | 26 | 37.9 | 63.4 | 62.9 | 68 |

**Sensivity study on dataset.**

State-of-the-art object detection models typically report their performance on the COCO dataset, a benchmark designed for general-purpose object detection. However, as shown in **Fig. 11**, the results on COCO and VisDrone differ significantly, highlighting the challenges of transferring models trained on COCO to domain-specific datasets like VisDrone. The VisDrone dataset, characterized by small, densely packed objects in complex aerial scenes, exposes limitations in models optimized for COCO. Among the evaluated models, YOLOv8 demonstrates strong performance on VisDrone, achieving competitive accuracy across various object sizes and maintaining robustness in detecting small and medium objects, which are critical for aerial tasks. This suggests that YOLOv8 is well-suited for domain-specific applications, particularly in aerial object detection scenarios.

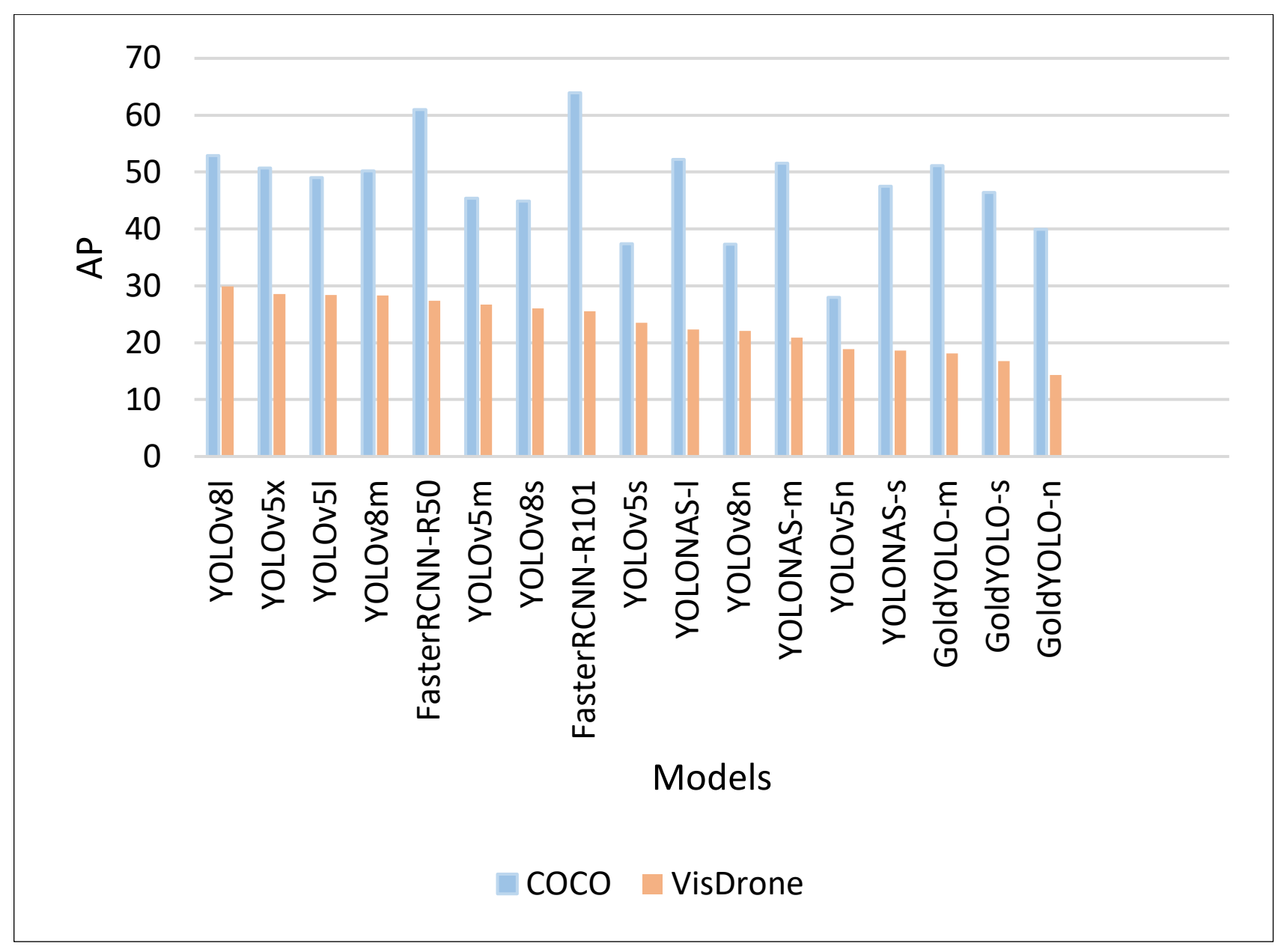

**Fig. 11.** Model performance comparison on COCO vs VisDrone dataset

**Sensivity study on α.**

We investigated several strategies for adjusting the distillation weight α, including constant, exponential decay, time-based decay, cosine annealing, and inverse sigmoid decay. As shown in **Table 13**, while some dynamic approaches marginally improved $AP_{large}$, none outperformed the fixed setting of α = 0.5 overall. These findings suggest that while dynamic α strategies offer theoretical flexibility, their practical effectiveness may depend on the dataset characteristics or stage of training. In our case, the constant α provided more consistent convergence behavior.

**Table 13.** Effect of dynamic and constant distillation α parameter adjustment

| Model | AP | AP50 | AP75 | $AP_{small}$ | $AP_{medium}$ | $AP_{large}$ |
|---|---|---|---|---|---|---|
| α=0.5 | 26.8 | 47.9 | 26 | 38.1 | 63.5 | 64.7 |
| Exponential decay | 26.7 | 47.4 | 26.3 | 37.6 | 63.1 | 67.3 |
| Time-based decay | 26.7 | 47.4 | 26.2 | 37.7 | 62.9 | 66.3 |
| Inverse sigmoid decay | 26.4 | 47.1 | 25.8 | 37.4 | 62.7 | 64.3 |
| Cosine annealing | 26.6 | 47.3 | 25.8 | 37.6 | 62.6 | 65.6 |

## 5 Conclusion

This paper introduces a novel model compression framework for YOLOv8, combining structured channel pruning with channel-wise knowledge distillation (CWD) to achieve an effective trade-off between accuracy and computational efficiency. To our knowledge, this is the first study to integrate CWD-based knowledge distillation with channel pruning on YOLOv8 for aerial object detection, enabling the retention of critical features despite aggressive compression.

We began by applying structured channel pruning to reduce the model's size and FLOPs. To counteract the potential degradation in detection performance, we employed CWD from the original unpruned YOLOv8m (teacher) to the compressed model (student), aligning their intermediate feature distributions to preserve important representational capacity.

Our proposed method achieves a 73.5% reduction in parameters (from 25.8M to 6.8M), a 73.2% drop in FLOPs, and a 65.8% reduction in MACs, while maintaining 47.9% AP50, only 2.7 lower than the base model. It also outperforms many popular lightweight object detectors, such as YOLOv5s, YOLOv8n, Gold-YOLO and GHOST in both accuracy and efficiency.

In terms of real-time performance, integrating TensorRT into our deployment pipeline improves inference speed from 26 FPS to 68 FPS, confirming the suitability of our approach for resource-constrained environments.

Future work may explore adaptive pruning strategies, more advanced distillation objectives, or integrating quantization techniques to further compress the model with minimal performance degradation. Additionally, expanding evaluation to more diverse datasets could further validate the generalizability of our approach.